\documentclass[runningheads]{llncs}
\usepackage[T1]{fontenc}
\usepackage{graphicx,verbatim}
\usepackage{amsmath,amssymb}
\usepackage{pifont}
\usepackage{multirow} 
\usepackage{booktabs} 
\usepackage{colortbl}
\usepackage[table,xcdraw]{xcolor} 
\usepackage{subcaption}

\DefineNamedColor{named}{ForestGreen} {cmyk}{0.91,0,0.88,0.12}

\begin{document}
\title{No Modality Left Behind: Dynamic Model Generation for Incomplete Medical Data}
\titlerunning{No Modality Left Behind}

\author{Christoph F\"{u}rb\"{o}ck\inst{1,2,3}
\and
Paul Weiser\inst{1,2,4,5}
\and
Branko Mitic\inst{1,2,3}
\and
Philipp Seeb\"{o}ck\inst{1,2,3}
\and
Thomas Helbich\inst{6}
\and
Georg Langs\inst{1,2,3}*
}
\authorrunning{C. F\"{u}rb\"{o}ck et al.}
%
\institute{
Computational Imaging Research Lab, Department for Biomedical Imaging and Image-guided Therapy, Medical University of Vienna, Vienna, Austria
\and
Comprehensive Center for Artificial Intelligence in Medicine, Medical University of Vienna, Vienna, Austria
\and
Christian Doppler Laboratory for Machine Learning Driven Precision Imaging, Department of Biomedical Imaging and Image-guided Therapy, Medical University of Vienna, Vienna, Austria
\and
Athinoula A. Martinos Center for Biomedical Imaging, Massachusetts General Hospital, Boston, Massachusetts USA
\and
Department of Radiology, Massachusetts General Hospital, Harvard Medical School, Boston, Massachusetts USA
\and
Division of General and Pediatric Radiology, Department of Biomedical Imaging and Image-guided Therapy, Medical University of Vienna, Vienna, Austria
}

\maketitle              
\begin{abstract}
In real world clinical environments, training and applying deep learning models on multi-modal medical imaging data often struggles with partially incomplete data. Standard approaches either discard missing samples, require imputation or repurpose dropout learning schemes, limiting robustness and generalizability. To address this, we propose a hypernetwork-based method that dynamically generates task-specific classification models conditioned on the set of available modalities. Instead of training a fixed model, a hypernetwork learns to predict the parameters of a task model adapted to available modalities, enabling training and inference on all samples, regardless of completeness. We compare this approach with (1) models trained only on complete data, (2) state of the art channel dropout methods, and (3) an imputation-based method, using artificially incomplete datasets to systematically analyze robustness to missing modalities. Results demonstrate superior adaptability of our method, outperforming state of the art approaches with an absolute increase in accuracy of up to 8\% when trained on a dataset with 25\% completeness (75\% of training data with missing modalities). By enabling a single model to generalize across all modality configurations, our approach provides an efficient solution for real-world multi-modal medical data analysis.


\end{abstract}
\let\thefootnote\relax\footnote{*Corresponding Author: georg.langs@meduniwien.ac.at}
\section{Introduction}

Deep learning has demonstrated remarkable success in medical image analysis, yet handling multi-modal data remains a significant challenge. Real-world data is often incomplete, missing modalities across individuals due to acquisition constraints, patient conditions, or cost~\cite{shadbahr2023impact,yang2023hyper}. Standard deep learning models struggle in such scenarios, as they either require complete data or rely on imputation techniques that may introduce bias and uncertainty~\cite{shadbahr2023impact}.

A prime example for multi-modal imaging is brain tumor analysis, where multi-parametric MRI scans provide complementary information about tumor biology (e.g., T1, T1ce, T2, FLAIR). In the BraTS 2019 dataset~\cite{menze2014multimodal,bakas2017advancing,bakas2018identifying}, multi-parametric MRI scans are used to segment cancer lesions and predict patient overall survival (OS) time. Several studies have explored OS prediction on the BraTS 2019 dataset, combining segmentation models with radiomic features and machine learning techniques~\cite{islam2020brain,wang2020automatic,agravat2019brain,wang20193d}. While these methods are effective in case of complete multi-modal data, they are not designed to handle missing modalities. In practice, not all modalities are always available, making robust multi-modal learning essential, and severily limiting approaches relying on complete data.

\begin{figure}[t]
    \centering
    \includegraphics[width=0.99\linewidth]{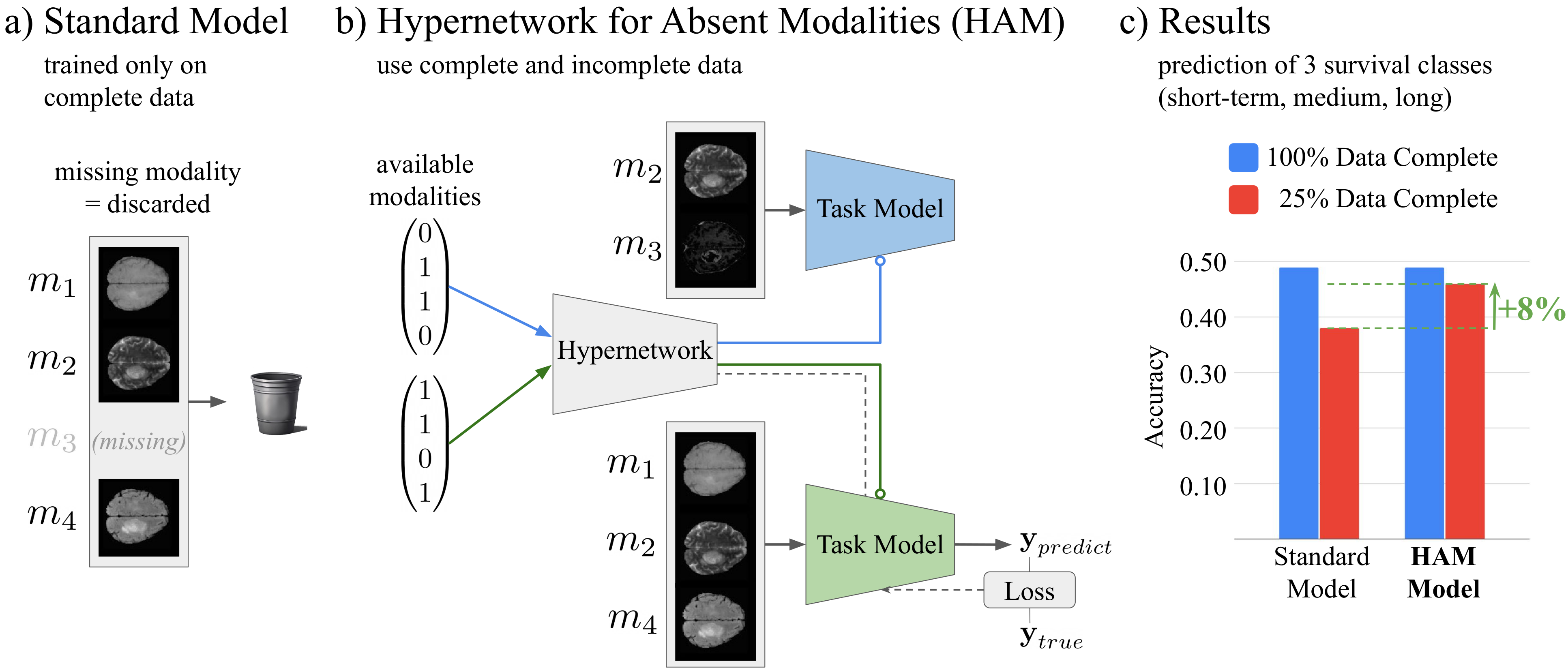}
    \caption{Overview of our hypernetwork-based approach for robust classification with absent modalities. a) Standard models discard samples with missing modalities during training. b) Going HAM utilizes all data by dynamically generating task-specific classifiers conditioned on available modalities, enabling inference across different modality subset examples, e.g., $(m_2, m_3)$ and $(m_1, m_2, m_4)$. c) When only 25\% of data samples have all modalities available, the HAM model outperforms the standard model by 8\% absolute accuracy.}
    \label{fig:graphical_abstract}
\end{figure}

\paragraph{Related work}
Existing methods to handle missing modalities can be categorized into three main strategies: (1) \textit{Complete data training} only uses samples for which all modalities are available (see Fig.~\ref{fig:graphical_abstract}a), ensuring high data quality but discarding part of the data. (2) \textit{Channel dropout} trains a model with randomly dropped input channels, e.g. \cite{neverova2015moddrop,zhao2023head,feng2023brain}, improving robustness but may come at the cost of a performance drop for complete data, since model parameters are not adapted to available modalities. (3) Imputation-based methods synthesize missing modalities using learned transformations, e.g. \cite{wang2023multi,ma2021smil,hamghalam2021modality,chartsias2017multimodal}. While effective, imputation can introduce artifacts and inconsistencies that degrade performance since they are based on prior assumptions of imputation models.

A notable recent approach by Yang and Sun \cite{yang2023hyper} employs a hypernetwork to perform imputation. It addresses missing modalities by using a generative model to reconstruct missing modalities, where the hypernetwork modulates the generative model using a modality code. 

\paragraph{Contribution}
We propose a novel approach to cope with partially incomplete multi-modal data during training and test time (Fig.~\ref{fig:graphical_abstract}b). We introduce a hypernetwork that learns to generate a task model (performing the downstream task, e.g. classification) adapted to available modalities. During training it uses all available data and does not rely on an imputation model as this may introduce bias or is limited by prior assumptions. During test time, it generates a task specific model that is optimized as a whole for the available set of modalities, instead of making a single model robust by dropout approaches. This offers the opportunity for a task model using its entire capacity for the present modalities, instead of a single task model that is robust to missing values. Results show that this strategy outperforms the state of the art.

\section{Hypernetwork for Absent Modalities (HAM)}\label{sec:methods}

We are given training examples $\langle \mathbf{I_i}, \boldsymbol{\mu}_i, y_i \rangle$, where \( \mathbf{I_i} \in \mathbb{R}^{m \times a \times b} \) is a multi-modal image with up to $m$ modalities, $a$ and $b$ are the spatial image dimensions, and $y_i$ is a task label. Missing modalities are padded with $0$ in $\mathbf{I_i}$, and $\boldsymbol{\mu}_i \in \{0,1\}^m$ indicates the presence of a valid modality in the example, where \( \mu_{j} = 1 \) if the \( j \)-th modality is present and \( 0 \) otherwise. We aim to train a task-specific model $f_ {\theta}: \mathbf{I_i} \mapsto y_i$ that can cope with missing modalities in $\mathbf{I_i}$.

To this end we introduce a hypernetwork that generates $\theta$ specifically for the available modalities in a sample.
To handle missing modalities, we go \textit{HAM} by training the hypernetwork $ h$ that predicts the task model's weights based on the available modalities. The hypernetwork maps the modality vector $\boldsymbol{\mu}$ to the corresponding task model parameters $\boldsymbol{\theta}$:
\begin{equation}
    h: \boldsymbol{\mu} \mapsto \boldsymbol{\theta}_{\boldsymbol{\mu}}
\end{equation}
The predicted weights \( \boldsymbol{\theta}_{\boldsymbol{\mu}} \) determine the task model $f$ enabling classification based on the available modalities:
\begin{equation}
    f_{\boldsymbol{\theta}_{\boldsymbol{\mu}}}: \langle\mathbf{I}_{i} \rangle_{\boldsymbol{\mu}} \mapsto y_i, \quad \text{where } \langle\mathbf{I}_{i} \rangle_{\boldsymbol{\mu}} \text{ contains only modalities with } \mu_j = 1.
\end{equation}
During training, the hypernetwork $h$ takes a randomly chosen $\boldsymbol{\mu}$ as input and generates the parameters $\boldsymbol{\theta}_{\boldsymbol{\mu}}$ of the classifier $f_{\boldsymbol{\theta}_{\boldsymbol{\mu}}}$ which then processes an input image $\langle\mathbf{I}_{i} \rangle_{\boldsymbol{\mu}}$ and produces predictions y$_{predict}$. The loss $\mathcal{L}($y$_{predict},$y$_{true})$ is calculated and backpropagation follows a two-stage process (see Fig.~\ref{fig:graphical_abstract}b): (1) the gradients of the loss are computed with respect to $\boldsymbol{\theta}_{\boldsymbol{\mu}}$, which would affect the weights generated by the hypernetwork; (2) these gradients are then propagated back through $h$, updating its parameters to improve the generation of task model weights. 

This approach ensures a single hypernetwork \( h \) is trained while dynamically generating task models \( f_{\boldsymbol{\theta}_{\boldsymbol{\mu}}} \) tailored to each specific modality configuration $\boldsymbol{\mu}$.

\section{Evaluation}

We evaluate the benefits of the proposed approach for both incomplete training- and test data. We use a publicly available brain cancer data set, and compare the classification accuracy with existing approaches to validate benefits across a range of missing data levels. 

\subsection{Dataset}
We utilize the \textbf{BraTS2019} dataset~\cite{menze2014multimodal,bakas2017advancing,bakas2018identifying} with multimodal 3D MRI scans of glioblastoma patients, corresponding segmentation masks and survival data. Each scan includes four MRI sequences: T1-weighted (T1), T1 with contrast enhancement (T1ce), T2-weighted (T2) and Fluid-Attenuated Inversion Recovery (FLAIR). 
Following prior work~\cite{wang2020automatic,asthana2022brain}, we categorize patient survival into three classes: \textit{Short-term survival} ($<$10 months), \textit{Medium-term survival} (10–15 months) and \textit{Long-term survival} ($>$15 months).  We split the official \textit{training set} into separate train (n=177) and test sets (n=36), consisting of 64/45/68 and 13/9/14 samples of these classes, respectively.
We pre-process the data by extracting a single 2D slice per scan, defined as the slice with the largest tumor area, and normalizing it to the range $[0,1]$.

\subsection{Experiments}

\paragraph{Experiment 'A':}
In a first experiment we evaluate the impact of different levels of missing modalities during training. To systematically evaluate performance under varying levels of missing modalities, we construct an artificially incomplete dataset. As the original dataset is complete, we randomly drop modalities for a subset of samples to simulate real-world missing data. We define a completeness level and randomly drop one modality per sample for all incomplete samples. The completeness level is defined as the percentage of complete samples in the dataset. For example, 75\% completeness means that 75\% of the dataset contains all modalities, while the remaining 25\% have exactly one missing modality. To assess the effectiveness of our approach and its dependence on missing data, we evaluate performance across multiple completeness rates and compare it to state of the art (SOTA) methods (trained on exactly the same dataset). Model performance is quantified using balanced accuracy and averaged over multiple runs. To test whether one method significantly outperforms another one, the Wilcoxon signed rank test is used with a significance level $\alpha = 0.05$.

\paragraph{Experiment 'B':} In a second experiment we evaluate the model with multiple missing modalities during training, and measure the impact of missing modalities during test time. We randomly drop $[1,m-1]$ modalities, i.e. at least one and up to $m-1$ at a fixed completeness level (25\%). This way we can use different missing modality sets during test-time, observe the performance of HAM and compare it to the SOTA methods (trained on exactly the same dataset).

\subsection{Comparative evaluation}

We compare the proposed approach with three SOTA methods. \textbf{(1) Complete Data Training:} A classification model trained exclusively on fully complete samples, i.e., only data points where all modalities are available are used for training. This model is defined as \textit{Standard}: $\mathbf{I}_i \mapsto y_i$, where $ \mu_{j} = 1 \quad \forall j$. While this ensures training on fully informative inputs, it discards a significant portion of available data. \textbf{(2) Channel Dropout Training:} A classification model trained on all available samples by randomly masking input modalities \textit{Dropout}: $\mathbf{I}_i \mapsto y_i$. Specifically, during training, a random modality vector \( \boldsymbol{\mu} \) is sampled, and a masking function is applied to set all channels corresponding to missing modalities (\(\mu_{j} = 0\)) to zero. \textbf{(3) Imputation-based Method:} By utilizing multiple encoders—one for each modality and one for a shared representation—both modality-specific and shared features are learned based on the work of Wang et al.~\cite{wang2023multi}. The features are then combined using residual feature fusion, imputing features of missing modalities, and passed to a decoder for final prediction \textit{FeatImpute}: $\mathbf{I}_i \mapsto y_i$ (see Appendix A).

\subsection{Implementation}

The hypernetwork responsible for generating task-specific model weights is implemented as a multi-layer perceptron (MLP), using \textit{ReLU} activations between layers (see Appendix B). The input to the hypernetwork is a binary modality vector $\boldsymbol{\mu}$, which indicates which modalities are available for each sample. The classification model used in this study---for the SOTA models and HAM---is based on the \textit{CBRTiny} architecture, a lightweight convolutional neural network tailored for medical image analysis~\cite{raghu2019transfusion}. It is designed to handle multi-modal input data, with adjustable input shape based on the available modalities.

The models are trained using a batch size $n_{batch}=16$ with a learning rate of $10^{-4}$. We employ weighted Focal Loss \cite{lin2017focal} to address class imbalance in the dataset, with the weights corresponding to the inverse ratio of the class frequencies to the total number of samples in the train set. For optimization, we use the \textit{RMSProp} algorithm for HAM, and \textit{Adam} is used for the SOTA methods. The stopping criterion, i.e. the number of iterations, for each model is defined by 5-fold cross-validation on the train set.

During training, we randomly select a binary modality vector $\boldsymbol{\mu}$  and the hypernetwork is trained to generate the corresponding task model weights based on this $\boldsymbol{\mu}$. Training proceeds for $n_{it}=10$ iterations before a new random $\boldsymbol{\mu}$ is selected as we have noticed a reselection each iteration does not yield a stable convergence and increasing $n_{it}$ enables the model to adapt to different modality combinations throughout the training process (see Appendix C).

In addition to early stopping, random augmentations are applied to the images to avoid overfitting, including random flipping as well as \textit{RandGaussianNoise, RandGaussianSmooth, RandAdjustContrast} and \textit{RandHistogramShift} from the MONAI library~\cite{monai}.

Experiments were conducted on one Nvidia RTX 2080 GPU (12GB VRAM).

\begin{table}[t]
    \caption{(a) Balanced accuracy after training with only part of the training data being complete (\%). (b) Sensitivity (Sens), specificity (Spec), and precision (Prec) averaged over all classes. ($^\ast$) indicates better than \textit{Standard}, and ($^\circledast$) better than \textit{Dropout} and \textit{FeatImpute} (if significant with $p<0.05$). }
    \label{tab:accuracy}
    \begin{subtable}[h]{0.53\textwidth}
        \centering
        \begin{tabular}{l>{\centering\arraybackslash}p{1.05cm}>{\centering\arraybackslash}p{1.05cm}>{\centering\arraybackslash}p{1.05cm}>{\centering\arraybackslash}p{1.05cm}}
            \multicolumn{1}{l}{\cellcolor{white} {(a)}} & \multicolumn{4}{c}{\textbf{Dataset Completeness}} \\ 
            \textbf{Model} & \textbf{100\%} & \textbf{75\%} & \textbf{50\%} & \textbf{25\%} \\
            \hline
            \textit{Standard} & \textbf{0.49} & \textbf{0.47} & 0.40 & 0.38 \\
            \rowcolor{gray!10} \textit{Dropout} & 0.45 & 0.45 & 0.43$^\ast$ & 0.44$^\ast$ \\
            \textit{FeatImpute} & 0.40 & 0.43 & 0.44$^\ast$ & 0.44$^\ast$\\
            \rowcolor{gray!10} \textbf{\textit{HAM}} & 0.49$^\circledast$ & 0.47 & \textbf{0.46}$^{\ast\circledast}$ & \textbf{0.46}$^{\ast\circledast}$ \\
        \end{tabular}
    \end{subtable}
    \hfill
    \begin{subtable}[h]{0.43\textwidth}
        \centering
        \begin{tabular}{l>{\centering\arraybackslash}p{1.05cm}>{\centering\arraybackslash}p{1.05cm}>{\centering\arraybackslash}p{1.05cm}}
        {(b)}&\multicolumn{3}{c}{\textbf{25\% Complete}}\\
            \textbf{Model} & \textbf{Sens} &  \textbf{Spec} & \textbf{Prec} \\
            \hline
            \textit{Standard} & 0.38 & 0.70 & 0.38  \\
            \rowcolor{gray!10} \textit{Dropout} & 0.44$^\ast$ & 0.73$^\ast$ & 0.44$^\ast$  \\
            \textit{FeatImpute} &  0.44$^\ast$ & 0.73$^\ast$ & 0.44$^\ast$ \\
            \rowcolor{gray!10} \textbf{\textit{HAM}} & \textbf{0.46}$^{\ast\circledast}$ & \textbf{0.74}$^\ast$ & \textbf{0.45}$^\ast$  \\
        \end{tabular}
    \end{subtable}
\end{table}
\begin{figure}
     \centering
     \begin{subfigure}[b]{0.49\linewidth}
         \centering
         \includegraphics[width=\linewidth,trim={0 70 0 0},clip]{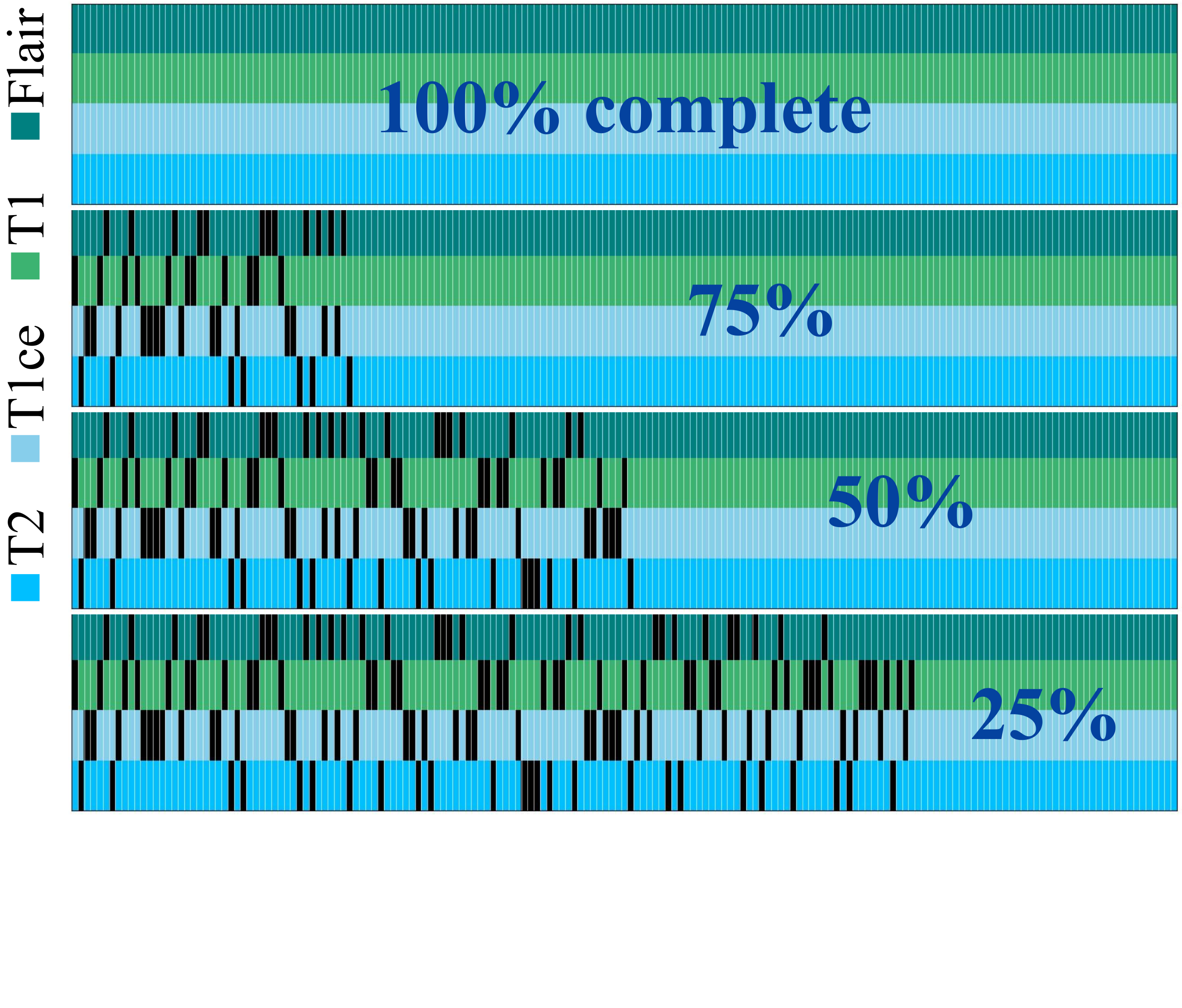}
         \caption{}
         \label{fig:incomplete_data_vis}
     \end{subfigure}
     \hfill
     \begin{subfigure}[b]{0.5\linewidth}
         \centering
         \includegraphics[width=\linewidth,trim={0 0 80 110},clip]{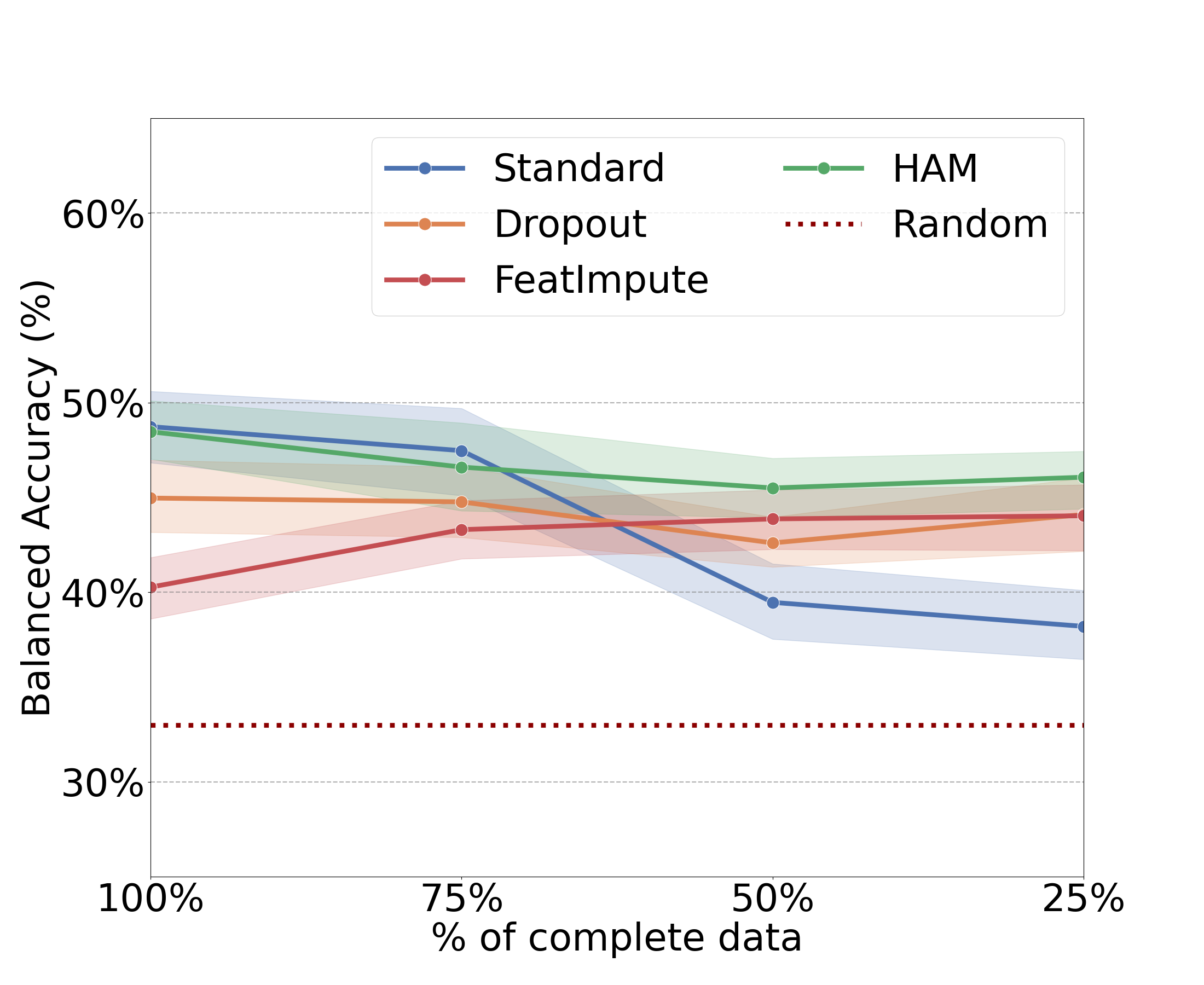}
         \caption{}
         \label{fig:accuracy_drop}
     \end{subfigure}
    \caption{(a) Visualization of the missing modalities for the different dataset completeness rates, and (b) the accuracy degradation as a function of data completeness rate. To evaluate the consistency of the results, models were trained 30 times, and the 95\% confidence interval is highlighted.}
    \label{fig:data_and_accu_drop}
\end{figure}

\section{Results}\label{sec:results}

\paragraph{Experiment 'A'}
Trained on the complete dataset (100\%, Fig\,\ref{fig:accuracy_drop}), \textit{HAM} achieves a balanced accuracy (BA) of 0.49, comparable to state-of-the-art methods relying on complete data~\cite{islam2020brain,wang2020automatic,agravat2019brain}, but better than methods such as Dropout (BA=0.45). As the ratio of training cases with all modalities decreases, \textit{HAM} stays comparatively stable (BA=0.46 at 25\%), while standard training drops to BA=0.38, and Dropout and Imputation yield BA=0.44. Table~\ref{tab:accuracy} summarizes measures of accuracy comparing \textit{HAM} with the state of the art approaches. Figure~\ref{fig:accuracy_drop} illustrates the accuracy degradation as the completeness rate decreases. 
The \textit{Dropout} model stays relatively stable, ranging from BA of 0.45 on the complete training data to 0.44 at 25\% completeness, but on a consistently lower level compared to \textit{HAM}.
At 25\% and 50\% complete training data, \textit{HAM} achieves better sensitivity, specificity, and precision compared to the \textit{Standard} model (Table~\ref{tab:accuracy}b). Compared to the \textit{Dropout} and \textit{FeatImpute} models, \textit{HAM} remains superior, with a statistically significant improvement in sensitivity, highlighting its robustness in distinguishing between classes even with highly incomplete data.
\paragraph{Experiment 'B'}
Based on a model trained on 25\% complete data, and multiple randomly removed modalities in the other training cases, we evaluated \textit{Dropout}, \textit{FeatImpute} and \textit{HAM} for different combinations of available modalities at test time (Table~\ref{tab:test_missing_modalities}). Unlike conventional multi-modal evaluations that primarily highlight performance improvements with increasing input modalities, our focus is on investigating the model's robustness and effectiveness regardless of the available modalities. The results show that \textit{HAM} consistently performs similar or better than \textit{Dropout} and \textit{FeatImpute} across all input scenarios, demonstrating its capability to handle missing modalities effectively. As expected, better performance is achieved when more modalities are available, highlighting the model's ability to integrate complementary information.
\begin{table}[t]
    \centering
    \caption{Testing with missing modality combinations. After training on data with 25\% completeness and one to three missing modalities per sample, the models are tested on different input modality combinations. Balanced accuracy (B. Accu) as well as sensitivity (Sens), specificity (Spec), and precision (Prec) averaged over all classes are given. ($^\ast$) indicates better than \textit{Dropout}, and ($^\circledast$) better than \textit{FeatImpute} (if significant with $p<0.05$).}
    \label{tab:test_missing_modalities}
    \begin{tabular}{>{\centering\arraybackslash}p{1.cm}>{\centering\arraybackslash}p{1.cm}>{\centering\arraybackslash}p{1.cm}>{\centering\arraybackslash}p{1.cm}ccccc}
        \textbf{T1} & \textbf{T2} & \textbf{T1ce} & \textbf{FLAIR} & \textbf{Model} & \textbf{B. Accu} & \textbf{Sens} & \textbf{Spec} & \textbf{Prec} \\
        \toprule
        \multirow{3}{*}{\textcolor{ForestGreen}{\checkmark}}  & \multirow{3}{*}{\textcolor{ForestGreen}{\checkmark}} & \multirow{3}{*}{\textcolor{red}{$\times$}} & \multirow{3}{*}{\textcolor{red}{$\times$}}   & \textit{Dropout} & 0.38 & 0.38 & 0.70 & 0.39  \\
         &  &  &    & \textit{FeatImpute}    & 0.38 & 0.38 & 0.70 & 0.39  \\
         &  &  &    & \textbf{\textit{HAM}}    & \textbf{0.38} & \textbf{0.38} & \textbf{0.70} & \textbf{0.40}  \\
        \arrayrulecolor{gray!30}\midrule
        \multirow{3}{*}{\textcolor{ForestGreen}{\checkmark}} & \multirow{3}{*}{\textcolor{red}{$\times$}} & \multirow{3}{*}{\textcolor{ForestGreen}{\checkmark}} & \multirow{3}{*}{\textcolor{red}{$\times$}}   & \textit{Dropout} & 0.39 & 0.39 & 0.70 & 0.42  \\
         &  &  &    & \textbf{\textit{FeatImpute}}    & \textbf{0.39} & \textbf{0.39} & \textbf{0.71} & \textbf{0.43}  \\
         &  &  &    & \textit{HAM}    & 0.38 & 0.38 & 0.69 & 0.41  \\
        \midrule
        \multirow{3}{*}{\textcolor{ForestGreen}{\checkmark}} & \multirow{3}{*}{\textcolor{red}{$\times$}} & \multirow{3}{*}{\textcolor{red}{$\times$}} & \multirow{3}{*}{\textcolor{ForestGreen}{\checkmark}}   & \textit{Dropout} & 0.40 & 0.40 & 0.71 & 0.42  \\
         &  &  &    & \textit{FeatImpute}    & 0.38 & 0.38 & 0.70 & 0.39  \\
         &  &  &    & \textbf{\textit{HAM}}    & \textbf{0.42}$^\circledast$ & \textbf{0.42}$^\circledast$ & \textbf{0.72}$^\circledast$ & \textbf{0.49}$^\ast$$^\circledast$  \\
        \midrule
        \multirow{3}{*}{\textcolor{red}{$\times$}} & \multirow{3}{*}{\textcolor{ForestGreen}{\checkmark}} & \multirow{3}{*}{\textcolor{red}{$\times$}} & \multirow{3}{*}{\textcolor{ForestGreen}{\checkmark}}   & \textit{Dropout} & 0.39 & 0.39 & 0.70 & 0.38  \\
         &  &  &    & \textit{FeatImpute}    & 0.39 & 0.39 & 0.70 & 0.38  \\
         &  &  &    & \textbf{\textit{HAM}}    & \textbf{0.42}$^\ast$$^\circledast$ & \textbf{0.42}$^\ast$$^\circledast$ & \textbf{0.72}$^\ast$$^\circledast$ & \textbf{0.43}$^\ast$$^\circledast$  \\
        \arrayrulecolor{gray}\midrule
        \multirow{3}{*}{\textcolor{ForestGreen}{\checkmark}} & \multirow{3}{*}{\textcolor{ForestGreen}{\checkmark}} & \multirow{3}{*}{\textcolor{ForestGreen}{\checkmark}} & \multirow{3}{*}{\textcolor{red}{$\times$}}  & \textit{Dropout} & 0.41 & 0.41 & 0.71 & 0.42  \\
         &  &  &    & \textit{FeatImpute}    & 0.40 & 0.40 & 0.71 & 0.41  \\
         &  &  &    & \textbf{\textit{HAM}}    & \textbf{0.41} & \textbf{0.41} & \textbf{0.72} & \textbf{0.41}  \\
        \arrayrulecolor{gray!30}\midrule
        \multirow{3}{*}{\textcolor{ForestGreen}{\checkmark}} & \multirow{3}{*}{\textcolor{ForestGreen}{\checkmark}} & \multirow{3}{*}{\textcolor{red}{$\times$}} & \multirow{3}{*}{\textcolor{ForestGreen}{\checkmark}}  & \textit{Dropout} & 0.41 & 0.41 & 0.71 & 0.43  \\
         &  &  &    & \textit{FeatImpute}    & 0.40 & 0.40 & 0.71 & 0.40  \\
         &  &  &    & \textbf{\textit{HAM}}    & \textbf{0.47}$^\ast$$^\circledast$ & \textbf{0.47}$^\ast$$^\circledast$ & \textbf{0.74}$^\ast$$^\circledast$ & \textbf{0.47}$^\ast$$^\circledast$  \\
        \arrayrulecolor{gray}\midrule
        \multirow{3}{*}{\textcolor{ForestGreen}{\checkmark}} & \multirow{3}{*}{\textcolor{ForestGreen}{\checkmark}} & \multirow{3}{*}{\textcolor{ForestGreen}{\checkmark}} & \multirow{3}{*}{\textcolor{ForestGreen}{\checkmark}} & \textit{Dropout} & 0.42 & 0.42 & 0.72 & 0.43 \\
         &  &  &    & \textit{FeatImpute}    & 0.42 & 0.42 & 0.72 & 0.43  \\
         &  &  &    & \textbf{\textit{HAM}}    & \textbf{0.46}$^\ast$$^\circledast$ & \textbf{0.46}$^\ast$$^\circledast$ & \textbf{0.73}$^\ast$ & \textbf{0.44}  \\
         \bottomrule
    \end{tabular}
\end{table}
\section{Discussion and Conclusion}\label{sec:discussion}

\paragraph{Summary of Key Findings}
In this study, we proposed a novel approach for handling missing modalities in multi-modal medical data using a hypernetwork. Instead of restricting training to complete examples, or training a single model that copes with missing data, \textit{HAM} generates a task-specific classification model adapted to available modalities, without the need for modality imputation.
Imputation methods attempt to estimate missing data, which can introduce biases and uncertainties, especially if the imputation model is not perfectly accurate~\cite{shadbahr2023impact}. \textit{HAM} bypasses the need for imputation by directly learning to handle varying modality combinations, thereby reducing the risk of propagating imputation errors through the model. 
Experimental results on the BraTS2019 dataset demonstrate that our approach outperforms standard methods, including training on complete data, channel dropout and feature imputation, in terms of classification accuracy when training on incomplete data. Specifically, our method's adaptability enables it to reach the upper-bound performance of the \textit{Standard} model when data is fully available, while also demonstrating significantly greater robustness to missing-modality scenarios compared to the SOTA methods.

\paragraph{Comparison to Related Work}
Compared to previous work on BraTS2019 survival prediction, \textit{HAM} offers a complementary benefit by handling missing modalities, a critical challenge in real-world clinical applications. Studies such as~\cite{islam2020brain,wang2020automatic,agravat2019brain} have demonstrated the effectiveness of combining deep learning-based segmentation with radiomic features, achieving accuracies between 0.43 and 0.55 when all modalities are available. Our model achieves a comparable 0.49 accuracy on complete data, reinforcing the strength of our approach in leveraging multi-modal information. However, unlike these methods, which assume fully available imaging data, our model is designed to handle incomplete inputs. Even when only 25\% of the training data is complete, we maintain a competitive accuracy of 0.46, while a model trained solely on complete data drops to 0.38. Compared to state of the art approaches for handling missing modalities, \textit{HAM} achieves significantly better performance in this setting. The escalating increase in accuracy of our approach over the SOTA methods as completeness decreases supports the hypothesis that generating weights specifically tailored to each modality configuration leads to more robust and reliable predictions. These results highlight that while existing methods perform well under ideal conditions, hypernetworks are a means to adapt network architectures to heterogeneous data availability, both during training and application to new data.

\paragraph{Limitations and Future Work}
Despite the promising results, one limitation is the relatively small size of the BraTS2019 dataset. At the same time, few other datasets offer such comprehensive image modality diversity. This makes BraTS2019 particularly well-suited for evaluating the robustness of our method to missing modalities. While we plan to extend the proposed approach to other tasks in future, such as tumor or organ segmentation, these extensions would not alter the underlying framework or methodology proposed. Instead, they would simply adapt the approach to new contexts, reinforcing its versatility. By effectively demonstrating the capabilities of our proposed method, this study paves the way for future research to explore its broader applicability.

\begin{credits}
\subsubsection{\ackname} This study was funded by the Vienna Science and Technology Fund (WWTF, PREDICTOME) [10.47379/LS20065], the European Commission under Grant Agreement No.101080302 AI-POD, and No.101219312 BreastSCan, and received co-funding from the European Union’s Horizon Europe research and innovation programme under grant agreement No.101100633 EUCAIM. The financial support by the Austrian Federal Ministry for Digital and Economic Affairs, the National Foundation for Research, Technology and Development and the Christian Doppler Research Association is gratefully acknowledged.

\subsubsection{\discintname}
The authors have no competing interests to declare that are
relevant to the content of this article.
\end{credits}

%
%
%
\bibliographystyle{splncs04}

%

\section*{Appendix}

\subsection*{Appendix A: Feature Imputation using Shared-Specific Representation Learning}

To enable a state-of-the-art performance baseline of an imputation based method, we implemented \textit{FeatImpute} based on the ShaSpec framework~\cite{wang2023multi}, which separates and recombines modality-specific and shared features to support robust generalization under missing modalities.

The \textit{FeatImpute} architecture consists of the following key components:\\
\textbullet\, \textbf{Specific Encoders:} Four parallel encoders (\texttt{CBRTiny} without classification heads) extract modality-specific features, activated based on the modality mask $\boldsymbol{\mu}$ indicating availability of each modality.\\
\textbullet\,  \textbf{Shared Encoder:} A unified encoder (\texttt{CBRTiny} without classification heads) processes stacked inputs to produce shared representations, later disentangled per modality.\\
\textbullet\,  \textbf{Compositional Layer:} For each available modality, a \texttt{CompositionalLayer} fuses its specific and shared features. The features for missing modalities are imputed with the shared features from the first available modality.\\
\textbullet\,  \textbf{Shared Classifier:} A shared linear head predicts from fused features.\\
\textbullet\,  \textbf{Domain Classifier:} An auxiliary linear classifier encourages specific features to remain modality-specific via a domain discrimination loss.

The model is trained using the Adam optimizer with a learning rate of $10^{-4}$ and the overall loss function combines the following terms:

{1) Task Loss $\mathcal{L}_{\text{task}}$:} The focal loss~\cite{lin2017focal} is employed for the main classification task to address class imbalance.
{2) Shared Consistency Loss $\mathcal{L}_{\text{shared}}$:} An $L_1$ loss is applied across shared feature vectors of all available modality pairs, enforcing consistency among shared representations.
{3) Specific Feature Loss $\mathcal{L}_{\text{spec}}$:} A cross-entropy loss is used for the auxiliary domain classification task, promoting the disentanglement of modality-specific features.
The total loss is given by:
\[
\mathcal{L}_{\text{total}} = \mathcal{L}_{\text{task}} + \lambda_{\text{shared}} \cdot \mathcal{L}_{\text{shared}} + \lambda_{\text{spec}} \cdot \mathcal{L}_{\text{spec}}
\]
with $\lambda_{\text{shared}} = 0.1$ and $\lambda_{\text{spec}} = 0.02$, consistent with the original work~\cite{wang2023multi}.

During training, input batches are dynamically sampled using random $\boldsymbol{\mu}$ to simulate missing or partial modalities, enhancing the model’s robustness to incomplete inputs.

\subsection*{Appendix B: Hypernetwork Architecture}

We employ a fully-connected multilayer perceptron (MLP) as a hypernetwork to generate the weights of a target task model, conditioned on input modality configurations. It consists of four linear layers with ReLU activations and a bottleneck hidden size of 4, mapping the modality vector $\boldsymbol{\mu}$ (\texttt{in\_sz}) to the CBRTiny weight space (\texttt{out\_sz})~\cite{raghu2019transfusion}, with the output truncated to the required size. The narrow hidden layers limit representational capacity, but in the context of a hypernetwork predicting weights for a small model—and given the limited dataset size—this design can be sufficient. The bottleneck thus represents a controlled trade-off between model complexity and overfitting risk, rather than a fundamental limitation.

\subsection*{Appendix C: Model Convergence}

\begin{figure}
     \centering
     \begin{subfigure}[b]{0.62\linewidth}
         \centering
         \includegraphics[width=\linewidth,trim={20 0 70 50},clip]{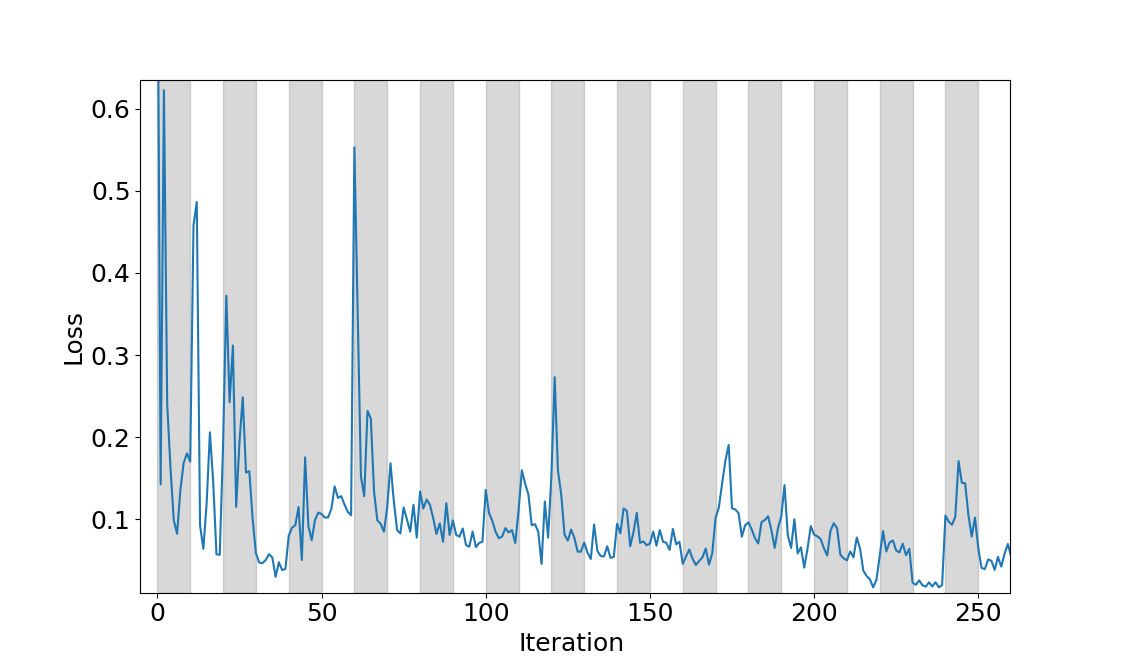}
         \caption{}
         \label{fig:loss_zoom}
     \end{subfigure}
     \hfill
     \begin{subfigure}[b]{0.35\linewidth}
         \centering
         \includegraphics[width=\linewidth,trim={0 0 0 20},clip]{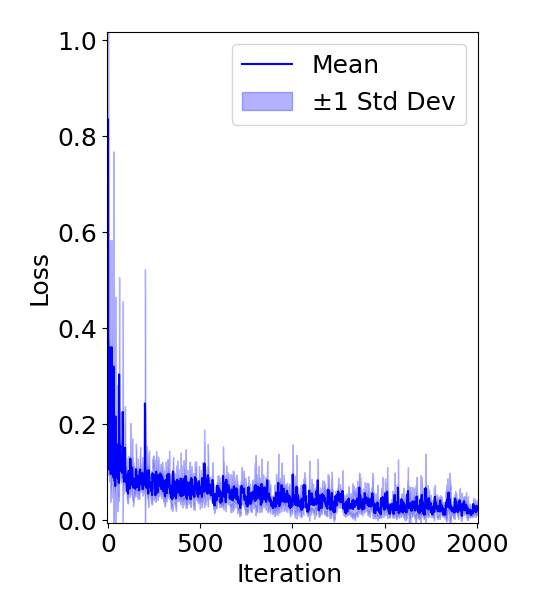}
         \caption{}
         \label{fig:loss_curve_avg}
     \end{subfigure}
    \caption{(a) Short-term effects of modality vector re-selection every ten iterations on training loss. (b) Long-term learning curve with standard deviation across 5 cross-validation runs.}
    \label{fig:loss_curves}
\end{figure}

An important hyperparameter in our training procedure is \( n_{\mathrm{it}} \), which controls the frequency at which a new modality vector \( \boldsymbol{\mu} \) is selected. Through empirical testing, we found that setting \( n_{\mathrm{it}} = 10 \) provides a good trade-off between stability and adaptability.
As shown in Figure~\ref{fig:loss_curves}a, re-selecting \( \boldsymbol{\mu} \) typically causes a temporary spike in the training loss, particularly in the early stages of training. This spike occurs because the hypernetwork has not encountered the newly sampled \( \boldsymbol{\mu} \) — that specific combination of modalities — during previous updates. As a result, the corresponding task network predicted by the hypernetwork initially performs poorly, yielding a higher loss. However, the hypernetwork adapts to each new configuration and as it is exposed to an increasing variety of \( \boldsymbol{\mu} \) values, its predictions improve across a broader range of modality combinations. This leads to a smoother and more stable training curve, ultimately allowing the model to converge and generalize effectively across the full space of \( \boldsymbol{\mu} \). Figure~\ref{fig:loss_curves}b illustrates the complete learning curve, averaged over five cross-validation runs (on the train set), along with the corresponding standard deviation.

\end{document}